\pdfoutput=1

\documentclass[10pt,twocolumn,letterpaper]{article}

\usepackage[pagenumbers]{cvpr} 

\usepackage{caption}
\usepackage{booktabs}
\usepackage{amsmath}
\usepackage{amssymb}
\usepackage{marvosym} 

\definecolor{cvprblue}{rgb}{0.21,0.49,0.74}
\usepackage[pagebackref,breaklinks,colorlinks,allcolors=cvprblue]{hyperref}

\def\paperID{*****} 
\def\confName{CVPR}
\def\confYear{2026}

\newcommand{\shortname}{ShotPlan}

\title{ShotPlan: Cinematic Video Generation with Learnable Planning Token}

\author{
Su Guo\textsuperscript{*,1,2} \quad
Guangce Liu\textsuperscript{*,1} \quad
Haosen Yang \quad
Jiepeng Wang\textsuperscript{1} \quad
Cong Liu\textsuperscript{1} \\
Junqi Liu\textsuperscript{1} \quad
Haibin Huang\textsuperscript{1} \quad
Hongxun Yao\textsuperscript{\Letter,2} \quad
Chi Zhang\textsuperscript{1} \quad
Xuelong Li\textsuperscript{\Letter,1} \\[6pt]
\textsuperscript{1}Institute of Artificial Intelligence (TeleAI), China Telecom \\
\textsuperscript{2}Harbin Institute of Technology \\[2pt]
{\tt\small guosu0930@gmail.com \quad h.yao@hit.edu.cn \quad xuelong\_li@chinatelecom.cn} \\[2pt]
{\small Project page: \url{https://pensioner-11.github.io/ShotPlan/}}
}

\begin{document}

\twocolumn[{%
\renewcommand\twocolumn[1][]{#1}%
\maketitle
\begin{center}
    \centering
    \captionsetup{type=figure}
    \includegraphics[width=\textwidth]{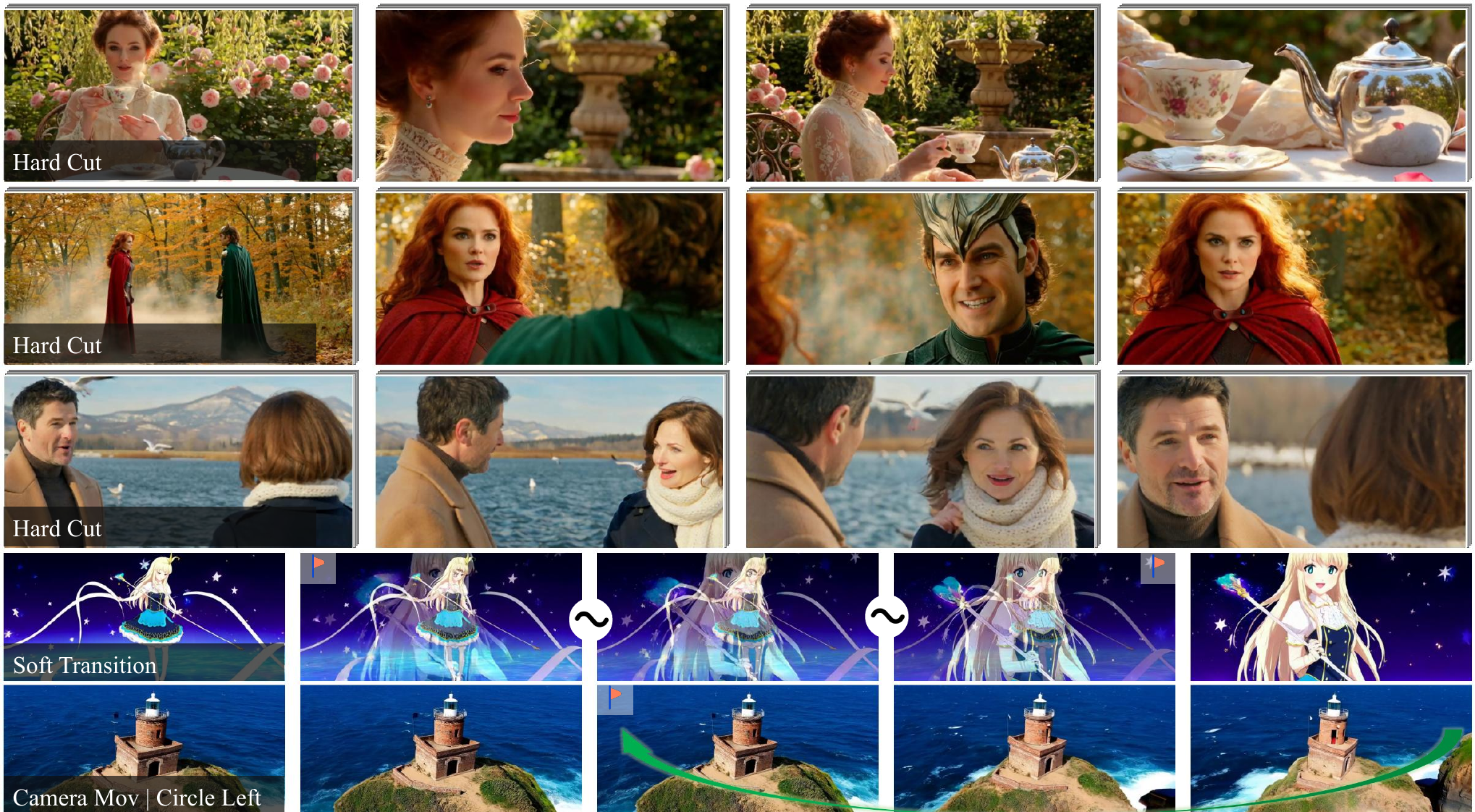}
    \captionof{figure}{
\shortname{} enables controllable multi-shot video generation within a single sequence.
(\textbf{Top}) Given user-specified timestamps, the model produces accurate hard cuts while preserving character identity and scene consistency across shots.
(\textbf{Middle}) The same framework also supports gradual transitions (e.g., cross-fades) in a unified manner.
(\textbf{Bottom}) As a byproduct, our approach can be extended to temporally localized camera motion control (e.g., circle-left), demonstrating its generality.
}
    \label{fig:teaser}
\end{center}%
}]

{\let\thefootnote\relax\footnotetext{\textsuperscript{*}Equal contribution. \quad \textsuperscript{\Letter}Corresponding authors.}}

\begin{abstract}
Current video generation models achieve impressive results in single-shot generation, yet remain limited in cinematic video generation, where coherent narratives and effective multi-shot composition require explicit shot planning. To address this challenge, we propose \textbf{\shortname}, a framework for explicit multi-shot cinematic video generation built upon a video diffusion foundation model. Our method introduces learnable planning tokens that capture shot-level transition cues and can be seamlessly integrated with the original video generation tokens to control transition timestamps. Unlike standard video generation tokens, the proposed planning tokens are equipped with Fractional Temporal Rotary Position Embedding (FRoPE), enabling shot transitions to be modeled at the frame level. Experiments demonstrate that \shortname{} significantly outperforms existing cinematic video generation methods, offering more flexible shot management and stronger inter-shot consistency.
\end{abstract}

\section{Introduction}
%
%
The video generation field has recently witnessed remarkable progress, as shown by
recent video diffusion models~\cite{ho2022video, singer2022make, blattmann2023stable, ho2022imagen, bar2024lumiere}.
This progress has been largely driven by Diffusion Transformers
~\cite{peebles2023scalable,zheng2024open,wu2025hunyuanvideo}, which are further scaled in recent
large-scale video models~\cite{yang2024cogvideox,wan2025wan}. These models have
enabled a broad range of AI-assisted video creation applications, producing
visually realistic single-shot videos.
However, these capabilities remain insufficient for practical video production settings, such as films, television series, and narrative storytelling, where creators require multi-shot video generation with explicit \emph{shot planning} over varying viewpoints, compositions, and scene arrangements. This planning enables creators to organize visual events temporally and guide the audience through coherent, structured narratives.

Straightforwardly, addressing this problem can be divided into image-to-video synthesis with keyframe generation~\cite{chen2023seine, xiao2025captain, zhao2024moviedreamer, zheng2024videogen}, where a sequence of keyframes is first produced by text-to-image models under consistency constraints, and each shot is then generated conditioned on its corresponding keyframe.
However, the effectiveness of such mechanisms remains limited, since individual shots are generated independently, preventing coherent inter-shot interaction during generation. Furthermore, overall video consistency relies heavily on the coverage and reliability of generated keyframes. When keyframes are sparse, they often fail to capture transient elements and maintain scene consistency across shots.

Beyond generating each shot independently, recent efforts~\cite{wu2025cinetrans,meng2025holocine,wang2025echoshot,wang2025multishotmaster} have explored generating cinematic video (i.e., multi-shot videos) directly within a single generation process. These methods typically introduce shot-planning mechanisms implicitly by modifying diffusion architectures, such as adding structured attention masks or RoPE-based positional discontinuities between shots. Although effective for shot separation, such designs can restrict cross-shot information interaction and introduce inconsistency. These architectural changes may also widen the gap between pretraining and finetuning, leading to suboptimal adaptation.


To address the aforementioned limitations, we introduce \textbf{\shortname{}}, a simple yet effective framework for multi-shot video generation. Our key insight is to explicitly represent shot transitions with learnable planning tokens, which serve as temporal anchors between shots. Each planning token is associated with a transition event and encodes shot-level cues to guide cross-shot generation. By integrating these tokens with the original video tokens, \shortname{} enables controllable transition timing while preserving the underlying video generation architecture.
A key challenge is that modern video diffusion models operate in temporally compressed latent spaces, where multiple physical frames correspond to a single latent timestep. As a result, transition signals aligned only to discrete latent indices cannot precisely match user-specified frame-level timestamps. We keep the original temporal Rotary Position Embedding (RoPE)~\cite{su2024roformer, wei2025videorope} for video tokens and introduce Fractional Temporal Rotary Position Embedding (FRoPE) for planning tokens, enabling shot transitions to be modeled at the frame level. This avoids relying on attention masks or manually altered positional distances to separate shots, while maintaining the original video generation architecture. As a result, \shortname{} expresses shot planning as a sequence of learnable tokens, enabling controllable multi-shot generation within a single coherent process.

Our contributions are:
(1) We introduce \textit{\shortname{}}, a simple yet effective framework that formulates multi-shot video generation as explicit shot planning within a single coherent generation process.
(2) To achieve this, we introduce learnable planning tokens and study their design, injection strategy, and Fractional Temporal Rotary Position Embedding (FRoPE), enabling frame-level shot-transition control while preserving the pretrained video generation architecture.
(3) Extensive experiments demonstrate that \shortname{} achieves controllable and coherent multi-shot generation, and further shows strong generalization potential for broader video control scenarios, such as camera movement control.

\section{Related Work}

\paragraph{Text-to-Video Generation.}
Diffusion-based models have become the dominant paradigm for text-to-video generation.
Early methods extend pretrained text-to-image models with temporal modules, motion layers, or spatio-temporal diffusion designs to synthesize short videos~\cite{guo2023animatediff, chen2023videocrafter1, chen2024videocrafter2, ho2022imagen, blattmann2023stable, bar2024lumiere}.
More recent advances adopt Transformer-based architectures, especially Diffusion Transformers (DiTs), to model video tokens in a scalable manner~\cite{peebles2023scalable, ma2024latte, ma2024sit}.
With large-scale video-text data and improved training pipelines, modern foundation video models have achieved substantial progress in visual fidelity, motion realism, text alignment, and temporal coherence~\cite{zheng2024open, yang2024cogvideox, wan2025wan, polyak2024movie}.
These models provide strong generative priors for video creation and have enabled a broad range of AI-assisted content production applications.
However, they are primarily optimized for continuous single-shot clips, where visual content evolves smoothly within one uninterrupted temporal segment.
In contrast, cinematic production often requires structured multi-shot videos with explicit shot boundaries, varying viewpoints, and coherent cross-shot semantics, which remains challenging for existing text-to-video models.

\paragraph{Multi-Shot Video Generation.}
Multi-shot video generation requires modeling discontinuous shot transitions while preserving subjects, scenes, and narrative consistency across shots.
One line of work decomposes the problem into keyframe or visual-plan generation followed by image-to-video synthesis~\cite{chen2023seine, zhou2024storydiffusion, he2023animate, xiao2025captain, zhao2024moviedreamer, zheng2024videogen,gong2023talecrafter, zhang2025storymem}.
Although intuitive, these pipelines generate shots largely independently and rely heavily on the quality and coverage of sparse keyframes, limiting inter-shot interaction during denoising.

Recent methods instead generate multi-shot videos within a unified denoising process.
ShotAdapter~\cite{kara2025shotadapter} introduces learnable transition tokens together with local attention masks to control shot number and duration.
CineTrans~\cite{wu2025cinetrans} uses attention masks to guide cinematic transitions, while HoloCine~\cite{meng2025holocine} and other long-context or mask-based methods~\cite{guo2025long, qi2025mask} design structured attention for multi-shot or multi-scene generation.
These methods improve shot separation, but their explicit attention constraints may limit cross-shot information propagation and weaken subject or scene consistency.

Another group of methods represents shot structure through temporal positional encoding.
RoPE~\cite{su2024roformer} is widely used in Transformer-based video generation, and recent work further studies video-oriented RoPE design~\cite{wei2025videorope}.
EchoShot~\cite{wang2025echoshot} and MultiShotMaster~\cite{wang2025multishotmaster} modify RoPE to distinguish shots or introduce shot-level phase offsets.
However, altering video-token positions changes their relative relationships and may disturb the pretrained temporal prior.
Moreover, VAE temporal compression makes frame-level shot boundaries difficult to align with discrete latent timesteps.

In contrast, we represent shot transitions as learnable planning tokens inserted into the video token sequence.
Unlike ShotAdapter, whose transition tokens are appended and depend on local attention masks, our planning tokens participate in standard self-attention without blocking cross-shot interactions.
We further equip them with Fractional Temporal RoPE while keeping the original RoPE of video tokens unchanged, enabling frame-level transition control without disrupting the pretrained positional structure.

\section{Method}
\label{sec:method}

\begin{figure*}[t]
    \centering
    \includegraphics[width=\textwidth]{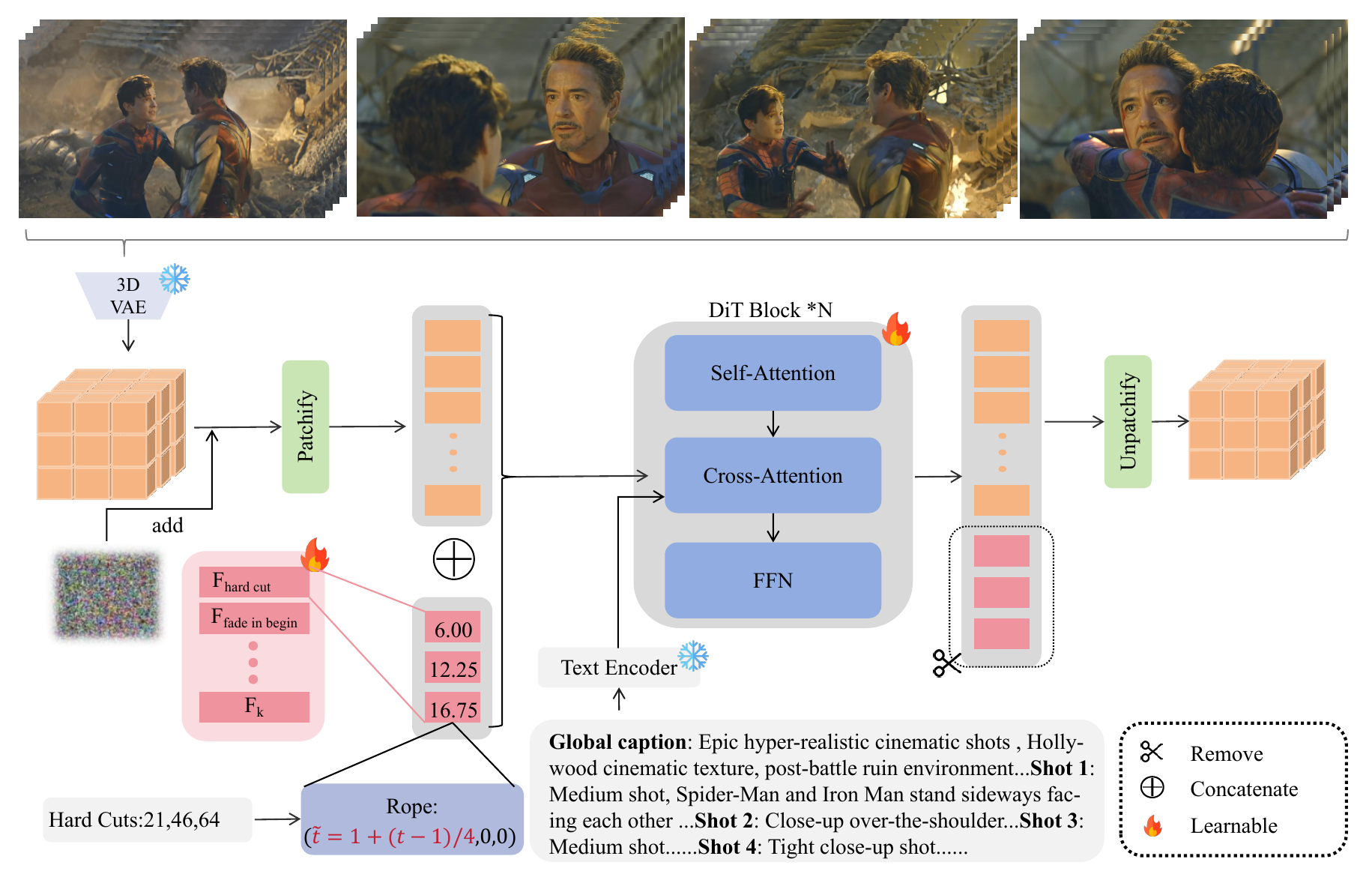}
    \caption{Overview of the \textbf{\shortname{}} framework. Given an input video, a 3D VAE encodes the frames into latent representations, which are then patchified into visual tokens. Based on user-specified cut timestamps and a structured prompt, \shortname{} introduces learnable planning tokens assigned with precise temporal coordinates via Fractional Rotary Position Embedding (FRoPE). These tokens are concatenated with visual tokens and jointly processed by DiT blocks, serving as temporal anchors for shot transitions. After denoising, the planning tokens are removed, and the updated visual tokens are unpatchified and decoded to generate the final multi-shot video.}
    \label{fig:pipeline}
\end{figure*}

In this work, we present \textit{\shortname}, a lightweight framework for multi-shot video generation with explicit shot planning within a single coherent generation process. Given a text prompt and user-specified transition timestamps, our goal is to generate coherent multi-shot videos in one pass. We first briefly review Diffusion Transformer (DiT)-based video diffusion models in Sec.~\ref{subsec:prelim}. Then, in Sec.~\ref{sec:method_flag_injection}, we introduce our \textit{Learnable Planning Token}, designed to capture and control shot-level transition cues. Next, Sec.~\ref{subsec:rope} presents Fractional Temporal Rotary Position Embedding (FRoPE) for these tokens. Finally, we describe our data curation pipeline for multi-shot training. An overview of \shortname\ is shown in Fig.~\ref{fig:pipeline}.


\subsection{Preliminaries}
\label{subsec:prelim}
Recent video generation models are commonly built on latent diffusion frameworks parameterized by Diffusion Transformers (DiTs)~\cite{peebles2023scalable, ma2024latte, wan2025wan}, and our method follows this paradigm. Given an input video $\mathcal{V} \in \mathbb{R}^{T \times H \times W \times 3}$ with $T$ frames, a pretrained 3D Variational Auto-Encoder (VAE)~\cite{yang2024cogvideox, wan2025wan} is used to reduce computational cost while preserving the semantic and perceptual content of the video. The video is compressed into a spatio-temporal latent representation:
\[
\mathbf{z}_0 = \mathcal{E}(\mathcal{V}), 
\quad 
\mathbf{z}_0 \in \mathbb{R}^{f \times h \times w \times C}.
\]

To enable transformer-based modeling, the latent tensor is further patchified into a sequence of tokens and processed by multi-head self-attention. Each token is associated with a 3D coordinate $(t,h,w)$, indicating its temporal and spatial location. Positional information is encoded using 3D Rotary Position Embedding (RoPE)~\cite{su2024roformer}, which applies axis-wise rotations to the query and key representations according to these coordinates. In particular, temporal positions are indexed by discrete latent steps, i.e., $t \in \{0,\dots,f-1\}$.

Following recent video diffusion models, we adopt the rectified flow formulation~\cite{lipman2022flow}. Given a clean latent $\mathbf{z}_0$ and Gaussian noise $\mathbf{z}_1 \sim \mathcal{N}(0,I)$, a linear probability path is defined as
\begin{equation}
\label{equation_addnoise}
\mathbf{z}_\tau = (1-\tau)\mathbf{z}_0 + \tau \mathbf{z}_1,
\quad
\tau \in [0,1].
\end{equation}
The DiT-based denoiser $f_\theta$ is trained to predict the target velocity with the following objective:
\begin{equation}
\label{eqation_loss}
\mathcal{L}_{flow}
=
\mathbb{E}_{\tau, \mathbf{z}_0, \mathbf{z}_1}
\left[
\left\|
\mathbf{v}_\theta(\mathbf{z}_\tau, \tau, c) - (\mathbf{z}_1 - \mathbf{z}_0)
\right\|_2^2
\right],
\end{equation}
where $c$ denotes the conditioning signal.
\subsection{Learnable Planning Token}
\label{sec:method_flag_injection}
Video Diffusion Transformers perform denoising over patchified latent tokens. Given an input video 
$\mathcal{V} \in \mathbb{R}^{T \times H \times W \times 3}$, 
we first obtain its clean latent representation 
$\mathbf{z}_0 \in \mathbb{R}^{f \times h \times w \times C}$. 
The noisy latent $\mathbf{z}_\tau$ is then obtained from $\mathbf{z}_0$ following Eq.~(\ref{equation_addnoise}) and patchified~\cite{peebles2023scalable, ma2024latte, ma2024sit} into a sequence of visual tokens:
\[
X_\tau = \mathrm{Patchify}(\mathbf{z}_\tau) \in \mathbb{R}^{N \times D}.
\]
To explicitly control shot transitions without modifying the attention structure or introducing positional discontinuities, we introduce learnable planning tokens as temporal anchors within the original coordinate system. 
Specifically, we define $F_{\text{p}} \in \mathbb{R}^{M \times D}$ and concatenate it with the patch-token sequence:
\begin{equation}
    X_{\text{in}} = [X_\tau \,\|\, F_{\text{p}}] 
    \in \mathbb{R}^{(N+M)\times D}.
\end{equation}
Here, $F_{\text{p}}$ is constructed from a small set of learnable token embeddings according to the transition type.
For hard cuts, we train a single learnable hard-cut token $\mathbf{f}_{\text{cut}} \in \mathbb{R}^{D}$: given $M$ user-specified cut points, this token is replicated $M$ times and concatenated with the video tokens, with each copy assigned the temporal coordinate of its own cut point via FRoPE (see Sec.~\ref{subsec:rope}) to distinguish their positions.
For gradual transitions (e.g., cross-fades), we instead train a start token and an end token, which are placed at the beginning and the end of the transition interval, respectively.
The concatenated token sequence is processed by the DiT backbone:
\[
X_{\text{out}} = \mathrm{DiT}_\theta(X_{\text{in}}, \tau, c),
\]
where $X_{\text{out}} \in \mathbb{R}^{(N+M)\times D}$. 
The planning tokens are jointly processed with the visual tokens across all transformer blocks, interacting with them through self-attention and serving as in-context conditioning~\cite{cai2025videocanvas} signals during denoising. 
Through training, $\mathrm{DiT}_\theta$ learns to adapt its predictions according to these temporally localized transition cues.

After denoising, we discard the planning tokens and retain the updated visual tokens 
$\tilde{X}_\tau \in \mathbb{R}^{N\times D}$. 
These tokens are then mapped back to the latent space via unpatchification:
\[
\hat{\mathbf{v}} = \mathrm{Unpatchify}(\tilde{X}_\tau),
\]
which corresponds to the predicted velocity 
$\mathbf{v}_\theta(\mathbf{z}_\tau, \tau, c)$ used for flow matching optimization.

\subsection{Fractional Temporal Rotary Position Embedding}
\label{subsec:rope}
\begin{figure*}[t]
    \centering
    \includegraphics[width=\textwidth]{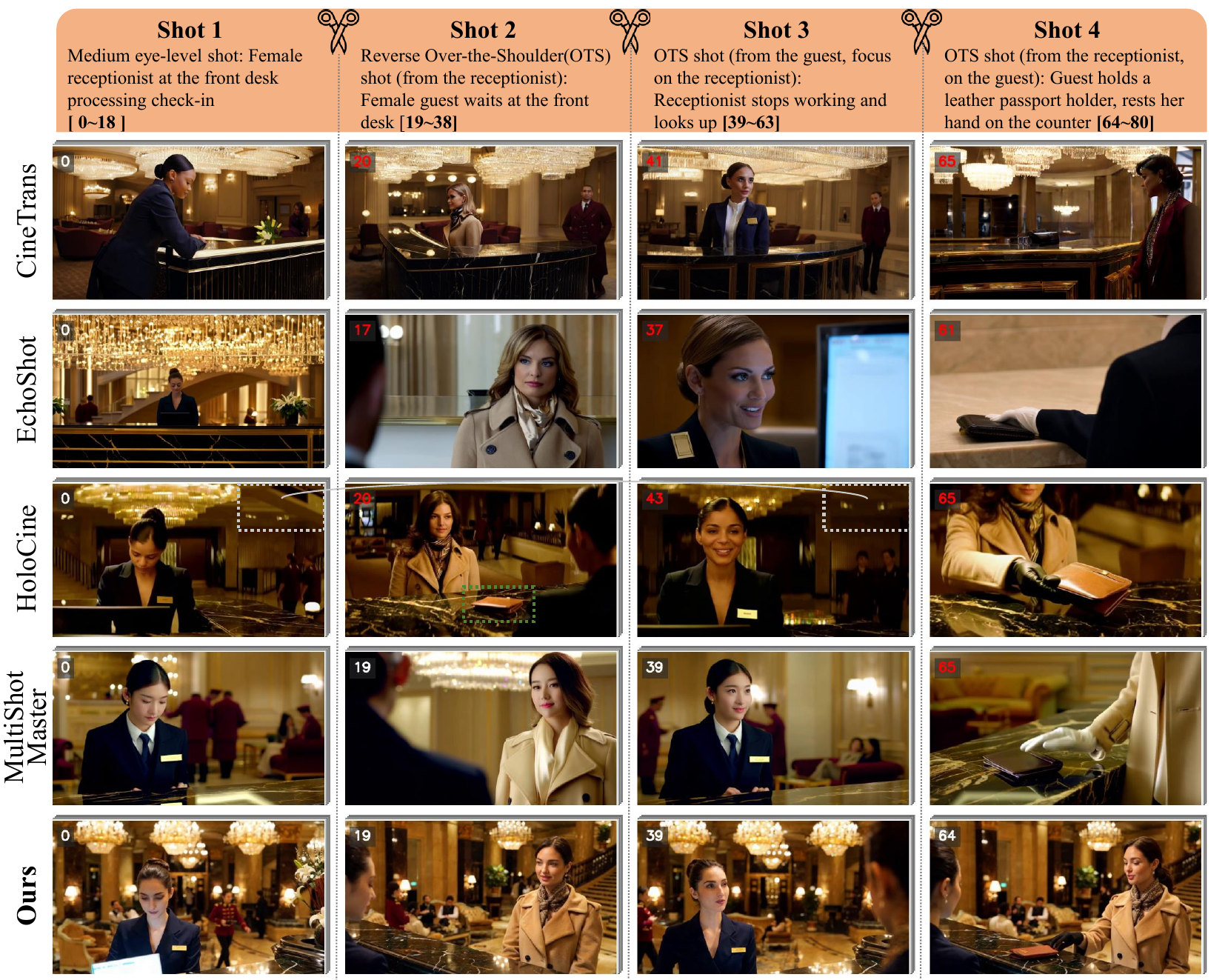}
    \caption{
        Qualitative comparison of multi-shot video generation with CineTrans~\cite{wu2025cinetrans}, EchoShot~\cite{wang2025echoshot}, HoloCine~\cite{meng2025holocine}, MultiShotMaster~\cite{wang2025multishotmaster} across a sequence of four continuous shots (Shot 1 to Shot 4). The number in the top-left corner of each frame indicates the starting frame index of each generated shot.
    }
    \label{fig:main_cmp}
\end{figure*}

We keep the original temporal RoPE for visual tokens and introduce Fractional Temporal Rotary Position Embedding (FRoPE) only for planning tokens. This allows precise frame-level transition control while preserving the pretrained positional encoding of the original video tokens.

Let the input video contain $T$ frames. With temporal compression ratio $s_t$, the latent temporal length is
\[
f = 1 + \frac{T - 1}{s_t}.
\]
Given a user-specified transition timestamp $t_u$ in the original frame space, we map it to a fractional latent coordinate:
\[
\tilde{t}_u =
\begin{cases}
0, & t_u = 0, \\
1 + \frac{t_u - 1}{s_t}, & t_u > 0,
\end{cases}
\]
where $\tilde{t}_u \in \mathbb{R}$.

Since RoPE is parameterized by sinusoidal functions~\cite{su2024roformer, wei2025videorope}, it naturally supports continuous coordinates without architectural changes. Therefore, FRoPE enables planning tokens to align with precise frame-level transition timestamps during generation instead of coarse latent indices. As planning tokens do not correspond to spatial patches, we assign them a fixed spatial coordinate $(h,w)=(0,0)$. The RoPE formulation for visual tokens remains unchanged. This preserves compatibility with the pretrained video diffusion backbone.

\subsection{Data Curation for Multi-Shot Training}
\label{subsec:datapipeline}
To construct training data with coherent multi-shot structures, we build our dataset starting from the open-source VideoEvent~\cite{liang2025videventlargedatasetunderstanding}, which consists of edited video clips with naturally dense shot transitions and strong narrative continuity.

We first apply standard preprocessing (e.g., subtitle and border removal), followed by TransNet V2~\cite{soucek2024transnet} to detect shot boundaries and segment each video into individual shots. 

To obtain training samples with sufficient temporal context, we adopt a sliding-window strategy over temporally contiguous shots. Specifically, we extract fixed-length windows of 5 seconds, each containing a sequence of consecutive shots. 
Each window contains at least two shots to ensure the presence of transition events, while each shot spans no fewer than 20 frames to maintain stable visual content.
Candidate windows are further filtered using Gemini 2.5~\cite{comanici2025gemini25pushingfrontier}, retaining only those that correspond to semantically consistent events with stable scene context and subject continuity.

For each selected sample, we further construct hierarchical annotations using Gemini. Each video is associated with a global description capturing overall scene attributes (e.g., environment, main subjects, lighting), along with per-shot captions that describe fine-grained details such as subject actions, camera viewpoints, and motion patterns. This hierarchical structure facilitates learning both cross-shot consistency and shot-specific variations during multi-shot video generation training.

\section{Experiment}

\subsection{Experimental Setup}
\paragraph{Implementation Details.}
We implement \shortname{} based on Wan2.1-T2V-14B~\cite{wan2025wan} at a resolution of $832 \times 480$ with 81 frames per sample. Videos are encoded using a 3D VAE with a temporal compression factor of 4. We train on over 7K hard-cut and 6K soft-cut samples for 3,500 steps using 8 NVIDIA H100 GPUs, with a batch size of 1 and a learning rate of $1\times10^{-5}$ using the AdamW optimizer. During training, both the planning tokens and DiT layers are jointly optimized. The planning tokens are initialized from a zero-mean Gaussian distribution with a standard deviation of 0.02. During inference, we use classifier-free guidance (CFG) with a scale of 5 and 50 denoising steps.
\paragraph{Baseline.}
We compare \shortname{} with representative multi-shot video synthesis methods, including CineTrans~\cite{wu2025cinetrans}, EchoShot~\cite{wang2025echoshot}, HoloCine~\cite{meng2025holocine}, and MultiShotMaster~\cite{wang2025multishotmaster}. These methods cover multi-shot narrative modeling, character-centric transitions, long-form cinematic generation, and recent multi-shot synthesis.

\paragraph{Evaluation.}
We construct 100 multi-shot prompts using Gemini 2.5~\cite{comanici2025gemini25pushingfrontier}, covering character consistency, scene consistency, and long-take scenarios. Specifically, 60 prompts focus on cross-shot character consistency, 30 on scene consistency, and 10 on long-take sequences without transitions. For each prompt, Gemini also specifies the shot pairs for evaluation, e.g., Shot 1 vs. Shot 3, enabling targeted assessment of cross-shot consistency. For fair comparison, we adapt each prompt to the input format required by each baseline.
\begin{table}[t]
\centering
\caption{Quantitative comparison across multiple metrics. Multi-shot generation is evaluated via automatic metrics, while camera movement is evaluated via a user study assessing motion type and timing accuracy.}
\label{tab:main_results}
\resizebox{\columnwidth}{!}{
\begin{tabular}{lccccc}
\toprule
       & Transition & Text & \multicolumn{2}{c}{Inter-shot Consistency} & Narrative \\
\cmidrule(lr){4-5}
Models & Deviation $\downarrow$ & Align $\uparrow$ & Character $\uparrow$ & Scene $\uparrow$ & Coherence $\uparrow$  \\
\midrule
\multicolumn{6}{c}{\textit{Multi-shot Generation}} \\
CineTrans        & 4.73          & 0.25          & 0.32          & 0.21          & 0.48          \\
EchoShot        & 6.95          & 0.26          & 0.31          & 0.23          & 0.51          \\
HoloCine        & 2.71          & \textbf{0.28} & 0.39          & 0.32          & 0.85          \\
MultiShotMaster       & 1.12          & 0.27          & 0.36          & 0.28          & 0.83          \\
Ours            & \textbf{0.64} & 0.26          & \textbf{0.46} & \textbf{0.37} & \textbf{0.88} \\
\bottomrule
\end{tabular}
}
\end{table}

\paragraph{Metrics.}
We evaluate multi-shot video generation from four aspects: 
(1) \textbf{Transition Deviation}, measured by the frame-level gap between detected transitions from TransNet V2~\cite{soucek2024transnet} and user-specified timestamps; 
(2) \textbf{Text Alignment}, computed as the ViCLIP~\cite{wang2023internvid} similarity between each generated shot and its shot-level caption; 
(3) \textbf{Cross-Shot Consistency}, measured by DINOv2~\cite{oquab2023dinov2} feature similarity between Gemini-specified shot pairs; 
and (4) \textbf{Narrative Coherence}, following MultiShotMaster~\cite{wang2025multishotmaster}, where Gemini 2.5 judges sampled frames and hierarchical captions across scene, subject, action, and spatial consistency.

\subsection{Quantitative Comparison}
Table~\ref{tab:main_results} reports the quantitative comparison on our multi-shot benchmark. 
Overall, \shortname{} achieves the best performance on most metrics. 
It substantially reduces transition deviation, achieving 0.64 compared with 1.12 from MultiShotMaster and 2.71 from HoloCine, demonstrating more precise shot-boundary control. 
For inter-shot consistency, \shortname{} also performs best, improving character consistency from 0.39 to 0.46 and scene consistency from 0.32 to 0.37 over the strongest baseline. 
Moreover, our method achieves the highest narrative coherence score of 0.88, outperforming HoloCine and MultiShotMaster. 
Although HoloCine obtains slightly higher text alignment, \shortname{} maintains competitive alignment while offering a better overall balance across temporal control, consistency, and coherence.

\subsection{Qualitative Comparison}
We further provide qualitative comparisons in Fig.~\ref{fig:main_cmp}.
The number in the top-left corner of each frame denotes the starting frame index of the corresponding generated shot. 
Compared with CineTrans~\cite{wu2025cinetrans} and HoloCine~\cite{meng2025holocine}, which use attention masking to control shot transitions, \shortname{} better preserves cross-shot information flow and scene structure. 
For example, in Row 3, HoloCine fails to preserve the background, where the staircase disappears in Shot 3. 
Although methods such as EchoShot~\cite{wang2025echoshot} and MultiShotMaster~\cite{wang2025multishotmaster} modify RoPE to model shot transitions, such positional manipulation may weaken spatial coherence across shots. 
As shown in Row 4, MultiShotMaster produces an inconsistent scene, where the lighting fixture on the left disappears in Shot 3. 
In contrast, \shortname{} introduces learnable planning tokens as temporal anchors, serving as in-context conditioning signals without modifying the attention structure or positional encoding. 
This allows our method to better preserve subject identity and scene layout across shots while accurately following user-specified transition timestamps.



\begin{table}[t]
\centering
\caption{Ablation study on token types, condition injection mechanisms, and positional encoding strategies.}
\label{tab:ablation}
\resizebox{\linewidth}{!}{
\begin{tabular}{l c c c c}
\toprule
Models & Transition & Text & \multicolumn{2}{c}{Inter-shot Consistency} \\
\cmidrule(lr){4-5}
       & Deviation $\downarrow$ & Align $\uparrow$ & Character $\uparrow$ & Scene $\uparrow$ \\
\midrule
\multicolumn{5}{c}{\textit{Token Design}} \\
Static Semantic Tokens  & 0.82 & 0.21 & 0.28 & 0.29 \\
\midrule
\multicolumn{5}{c}{\textit{Injection Mechanism}} \\
Latent Addition         & 1.76 &0.25 &0.42 & 0.36 \\
Global AdaLN            & 5.64 & \textbf{0.28} & 0.39 & 0.37 \\
Local AdaLN             & 4.39 & 0.26 & 0.37 & 0.36 \\
\midrule
\multicolumn{5}{c}{\textit{Positional Encoding}} \\
w/o Fractional RoPE     & 2.13 & 0.27 & 0.37 & 0.34 \\
\midrule
Ours (Full \shortname{}) &\textbf{0.64}&0.26&\textbf{0.46}  &\textbf{0.37}  \\
\bottomrule
\end{tabular}
}

\end{table}

\subsection{Ablation Study}
\label{sec:ablation}


\paragraph{Effectiveness of Learnable Tokens.}
We hypothesize that learnable planning tokens can capture shot-level transition cues through data-driven training. 
To validate this design, we compare them with static semantic tokens. 
Specifically, we extract the text embedding of the phrase ``scene cut'' using the T5 encoder, project it to the DiT token dimension through the text embedding layer, and apply average pooling to obtain a frozen token. 
This static token is then used to replace our learnable planning token during DiT fine-tuning.
As shown in Table~\ref{tab:ablation}, the non-learnable token leads to clear performance degradation, with lower character consistency, scene consistency, and text alignment. 
This suggests that a fixed semantic token is less compatible with the latent feature space of the pretrained DiT and forces the backbone to adapt to a rigid control signal. 
In contrast, learnable tokens can be jointly optimized with the model, better aligning with the underlying feature space and providing more effective conditioning for temporally localized shot control.
\paragraph{Analysis of Injection Mechanisms.}
To justify our in-context token injection strategy, we compare three alternative mechanisms for injecting learnable planning tokens. 
(1) \textbf{Latent Addition} directly adds the planning token to the visual token at the target transition timestamp. Although it provides basic control, it disrupts the interaction with RoPE and limits precision to coarse latent-level positions. 
(2) \textbf{Global AdaLN Modulation} maps the transition position to AdaLN modulation parameters via an auxiliary MLP. However, it provides only global conditioning and shows little improvement over the uncontrolled baseline. 
(3) \textbf{Local AdaLN Modulation} applies AdaLN only to the target latent position, improving localization but still lacking effective cooperation with RoPE, thus remaining limited to latent-level control. 

These comparisons show that treating transition control as in-context token modulation better preserves the pretrained DiT structure and enables more precise temporal control.

\paragraph{Effectiveness of Fractional RoPE.}
To evaluate the contribution of Fractional RoPE to precise temporal alignment, we replace the continuous fractional coordinates with standard discrete latent indices. 
As shown in Table~\ref{tab:ablation}, removing the fractional component degrades transition control from the physical-frame level to the coarser latent level. 
This confirms that Fractional RoPE is essential for bridging the granularity gap introduced by VAE temporal compression.
\begin{figure}
    \centering
    \includegraphics[width=\linewidth]{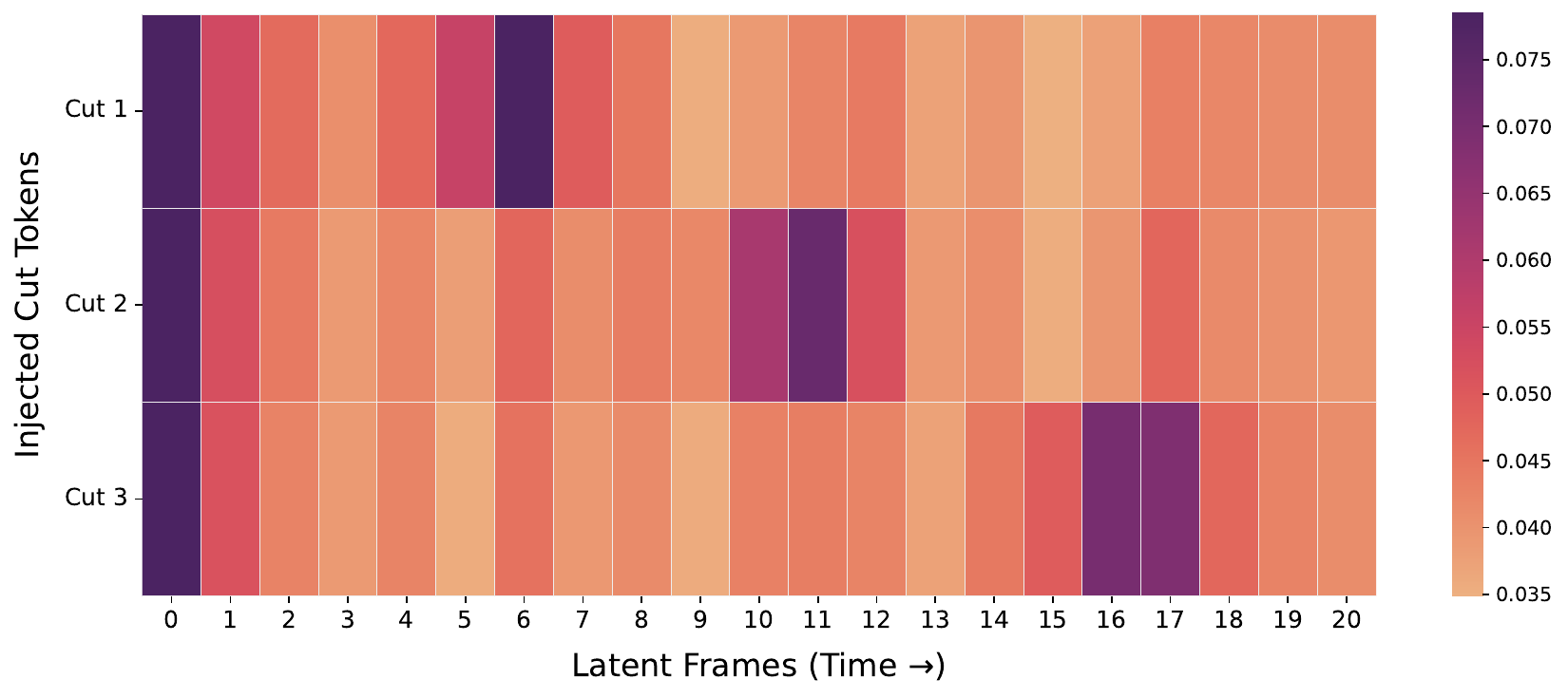}
    \caption{Latent temporal attention weight visualization of the injected cut tokens. The horizontal axis represents the temporal progression of latent frames, while the vertical axis represents the flag tokens corresponding to different cuts (Cut 1, Cut 2, Cut 3). }
    \label{fig:attention_map}
\end{figure}

\subsection{Attention Analysis of Planning Tokens}
To analyze how planning tokens regulate shot transitions, we visualize the temporal attention responses between planning tokens and video tokens from the 8-th DiT block, averaged across all attention heads. As shown in Fig.~\ref{fig:attention_map}, each planning token exhibits a clear attention concentration around its assigned cut timestamp, indicating that the model learns temporally localized transition control. Notably, the attention peak aligns with the projected fractional timestamp rather than discrete latent indices, validating the effectiveness of FRoPE for frame-level temporal alignment. The attention also spreads smoothly to neighboring frames, suggesting that the transition is achieved through soft temporal modulation instead of hard shot separation while preserving temporal semantic continuity across shots.

\begin{table}[t]
\centering
\caption{User Study on Camera Movement Control Accuracy.}
\label{tab:camera_movement}
\begin{tabular*}{\columnwidth}{@{\extracolsep{\fill}} lcc}
\toprule
 & \multicolumn{2}{c}{User Study} \\
\cmidrule(r){2-3}
Models & Type $\uparrow$ & Timing $\uparrow$ \\
\midrule
Wan 2.1 & 89\% & 97\% \\
Kling 2.6 & 81\% & 92\% \\
SeedDance 1.5 Pro & 86\% & 96\% \\
\bottomrule
\end{tabular*}
\end{table}

\begin{figure}
    \centering
    \includegraphics[width=\linewidth]{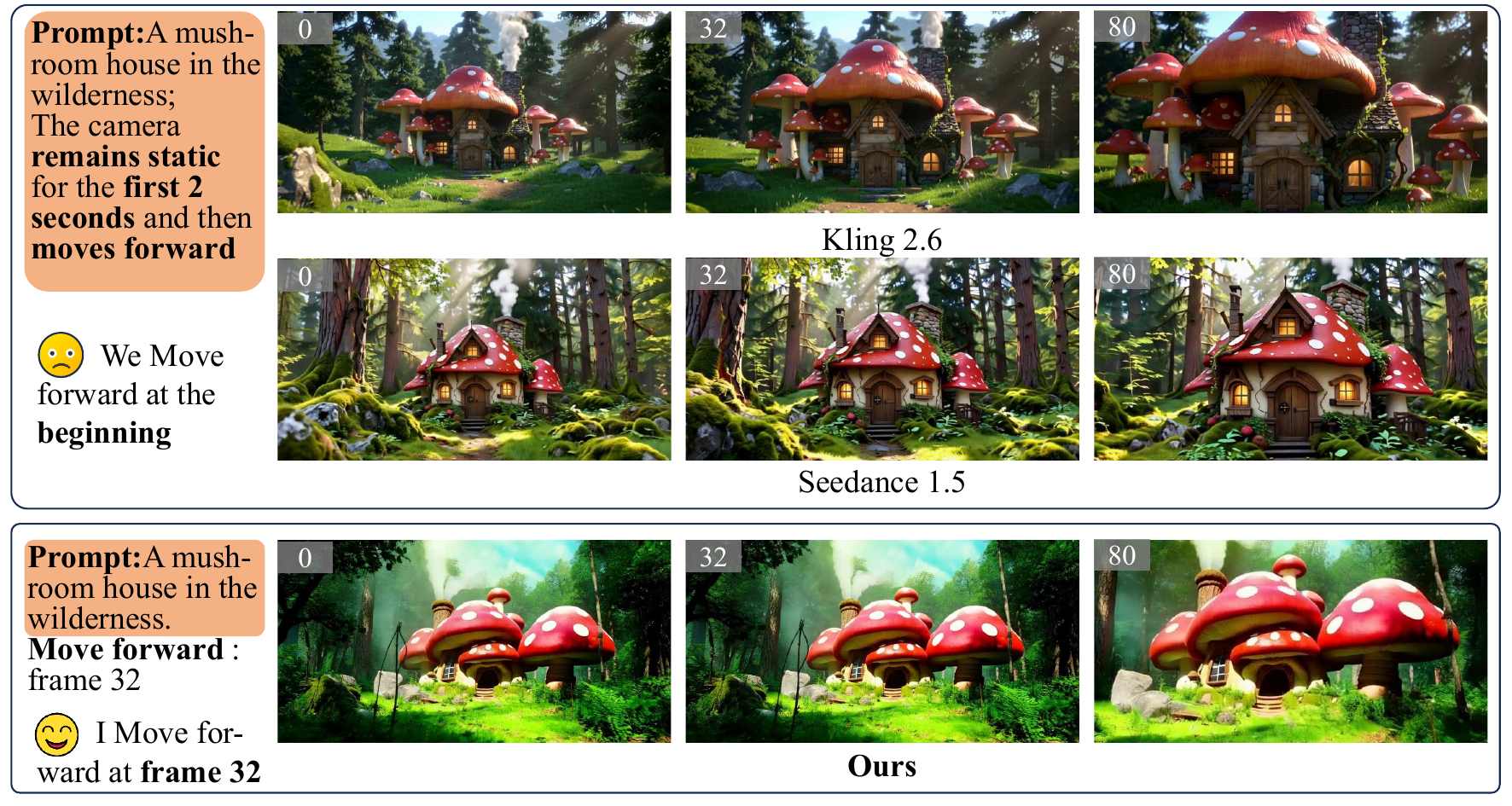}
    \caption{Qualitative comparison of temporally localized camera movement control. The figure compares ShotPlan (Ours) with commercial models (Kling 2.6~\cite{kling2024} and Seedance 1.5~\cite{seedance2025seedance}) in executing temporally localized camera motions.}
    \label{fig:camera_cmp}
\end{figure}
\subsection{Temporally Localized Camera Movement}
\label{subsec:camera_movement}

Our method extends beyond multi-shot generation and can be applied to temporally structured control tasks~\cite{he2024cameractrl, yang2024direct, wang2024motionctrl, zhao2024motiondirector, bai2025recammaster, geng2025motion}. 
As a representative example, we consider temporally localized camera movement, such as pan, zoom, or circular motion.

Unlike instantaneous shot transitions, camera motion evolves continuously over time. 
In our formulation, this can be handled with the same token mechanism: a control token is anchored at the user-specified starting timestamp, and the model learns to propagate its effect over subsequent frames. 
Conditioned on the corresponding motion semantics, the transformer generates coherent camera movement within the intended temporal interval without requiring explicit boundary tokens. 
As shown in Fig.~\ref{fig:camera_cmp}, \shortname{} accurately performs a ``circle left'' motion starting from the specified timestamp while maintaining smooth motion dynamics.

For quantitative evaluation, we conduct a user study comparing \shortname{} with Wan 2.1~\cite{wan2025wan} and commercial video generation systems, including Kling~2.6~\cite{kling2024} and SeedDance~1.5 Pro~\cite{seedance2025seedance}, as there are no public baselines for fine-grained temporally localized camera control. 
Participants evaluate two aspects: 
(1) \textbf{Motion Type}, whether the generated motion matches the textual instruction; and 
(2) \textbf{Timing Accuracy}, whether the motion starts at the specified timestamp and remains temporally localized without unintended drift. 
As reported in Table~\ref{tab:camera_movement}, \shortname{} achieves competitive motion-type accuracy and the highest timing accuracy, outperforming both commercial baselines. 
These results show that our learnable planning tokens, together with fractional temporal positioning, provide a unified mechanism for both discrete shot transitions and continuous temporal control.

\section{Conclusion}



We presented \shortname{}, a lightweight framework for explicit multi-shot cinematic video generation built upon a pretrained video diffusion model. By introducing learnable planning tokens with Fractional Temporal RoPE, \shortname{} enables frame-level control of shot transitions without modifying the original attention structure or positional encoding of the video backbone. The planning tokens serve as temporal anchors and in-context conditioning signals, allowing the model to manage shot boundaries while preserving the pretrained generative prior. Extensive experiments show that \shortname{} improves transition accuracy, inter-shot consistency, and narrative coherence over existing multi-shot video generation methods. We further demonstrate that the same formulation can be extended to temporally localized camera movement, suggesting its potential as a unified mechanism for structured temporal control in video generation.

{
    \small
    \bibliographystyle{ieeenat_fullname}
    \bibliography{references}
}

\clearpage
\appendix
\maketitlesupplementary
This supplementary material provides additional details and results of \shortname{}. Sec.~\ref{sec:gallery} presents a generation gallery of multi-shot cinematic videos. Sec.~\ref{sec:camera_data} details the construction of the camera-motion dataset used for the temporally localized camera movement experiments (Sec.~\ref{subsec:camera_movement} of the main paper). Sec.~\ref{sec:noise_ablation} provides an additional ablation on whether the planning tokens should be noised and included in the training loss.

\section{Generation Gallery}
\label{sec:gallery}

\begin{figure*}[t]
    \centering
    \includegraphics[width=1\linewidth]{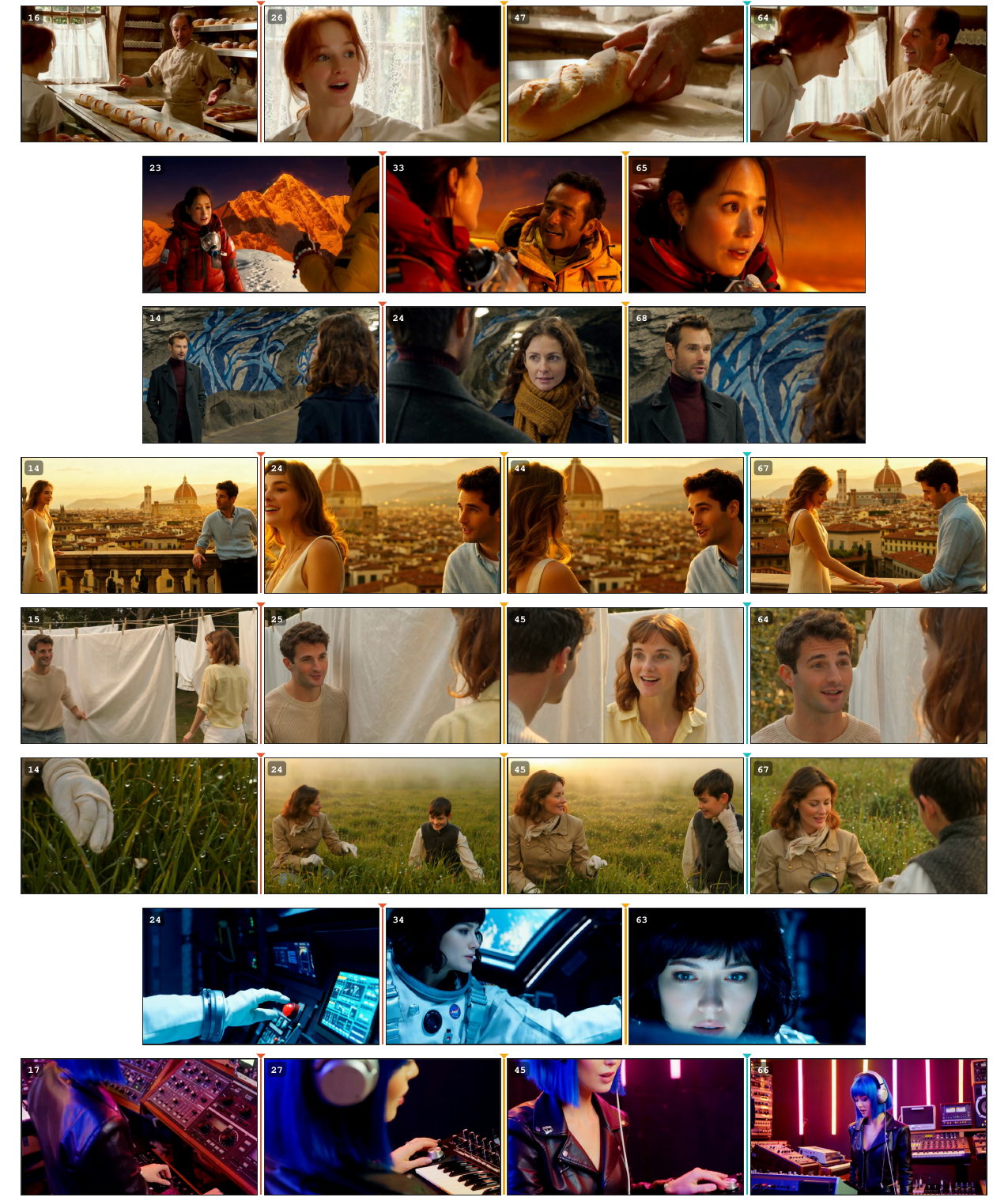}
    \caption{
\shortname{} generation gallery. Each row shows a multi-shot video generated by \shortname{}, with user-specified shot transitions.
}
    \label{fig:gallery1}
\end{figure*}

\begin{figure*}[t]
    \centering
    \includegraphics[width=1\linewidth]{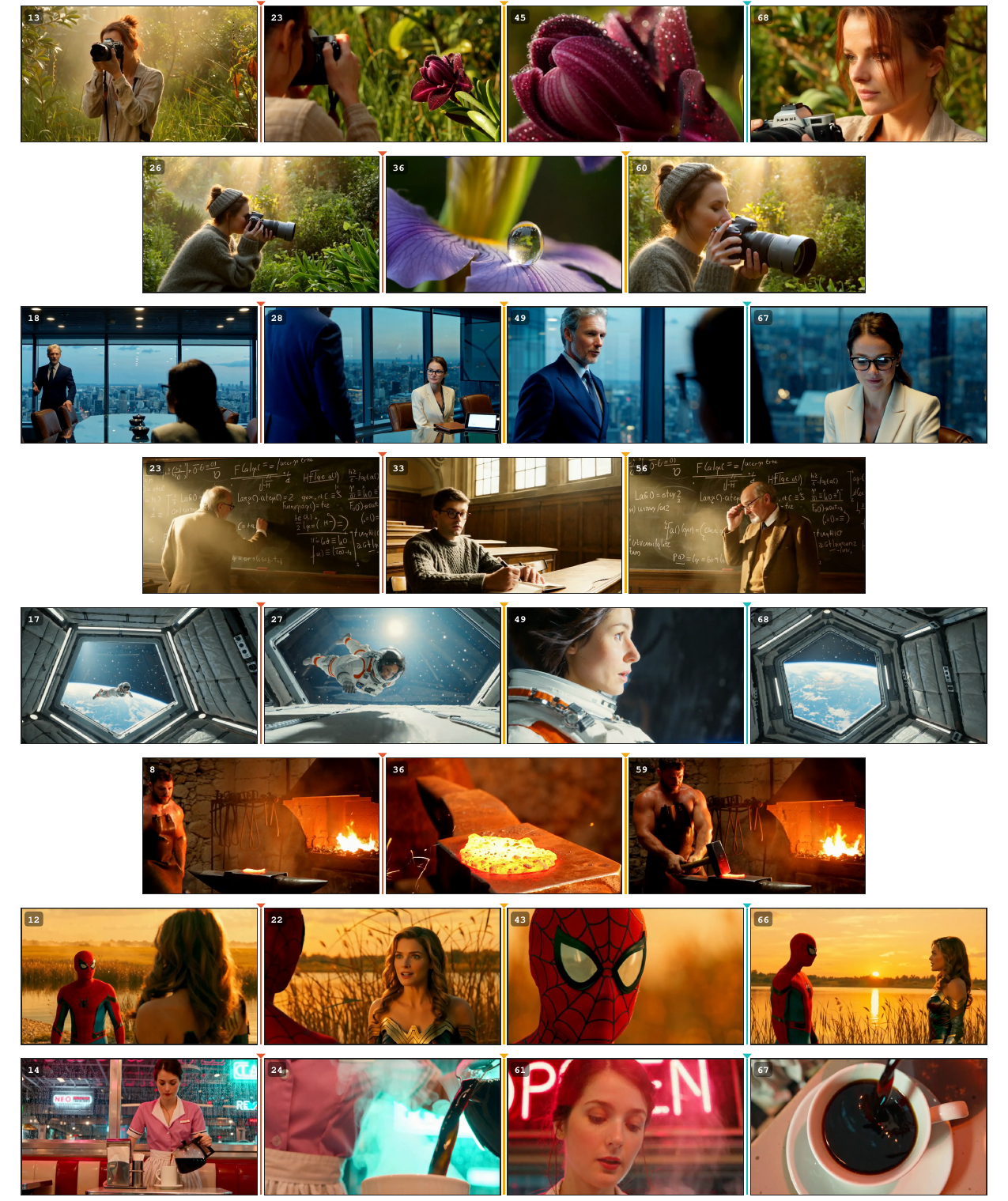}
    \caption{
\shortname{} generation gallery (continued). Each row shows a multi-shot video generated by \shortname{}, with user-specified shot transitions.
}
    \label{fig:gallery2}
\end{figure*}

Figures~\ref{fig:gallery1} and \ref{fig:gallery2} present a generation gallery of multi-shot cinematic videos produced by \shortname{}, illustrating diverse scenes, characters, and shot transitions with consistent subject identity and scene layout across shots.

\section{Camera-Motion Data Construction}
\label{sec:camera_data}

To support temporally localized camera movement control, we require training videos in which a prescribed camera motion begins at a known intermediate timestamp, while the preceding segment remains free of that motion. Since natural videos rarely provide such annotations, we synthesize the data as follows.

Given a source video, we first sample a random truncation point. The segment before the truncation point is kept unchanged, and the frame at the truncation point serves as the conditioning image for an image-to-video (I2V) model---Wan2.2~\cite{wan2025wan} fine-tuned with camera-motion LoRA~\cite{hu2022lora} adapters---which re-generates the remainder of the video such that it exhibits one of six predefined camera movements: circle-left, circle-right, move-left, move-right, zoom-in, and zoom-out. The original preceding segment and the generated segment are then concatenated, yielding a training sample whose camera motion starts exactly at the truncation point.

However, this naive concatenation introduces two typical failure modes: (1) a visible seam at the concatenation boundary, i.e., temporal discontinuity between the last original frame and the first generated frame; and (2) degraded scene consistency within the generated segment, since I2V generation conditioned on a single frame does not always preserve the original scene content. We therefore manually screen all concatenated videos and discard samples exhibiting either artifact. After filtering, we obtain approximately 6,000 videos for each camera-motion type.

\section{Ablation on Noising the Planning Tokens}
\label{sec:noise_ablation}

In our final design, the planning tokens are kept clean throughout training: they are neither perturbed by the diffusion noise nor included in the training loss. A natural concern is that inserting clean tokens into an otherwise noised token sequence makes the DiT input no longer follow a homogeneous noise distribution---part of the sequence is noised while the other part is not. In practice, this design is compatible with the flow-matching objective~\cite{lipman2022flow}, since the training loss is applied only to the video tokens while the planning tokens serve as deterministic conditioning anchors. This is conceptually related to image-conditioned video generation~\cite{wan2025wan}, where clean conditioning latents may be mixed with noised video latents. Different from image latents, our planning tokens are abstract learnable tokens that encode temporal control signals.

We also explored a noised variant in which the planning tokens are perturbed together with the video tokens and optimized using their detached values from the previous iteration as regression targets, with a 1,000-step learning-rate warmup. Under this training configuration, the variant converges substantially more slowly, requiring about 11,000 training steps to reach the performance achieved by the clean-token design within 3,500 steps. We therefore adopt the clean-token design for its simplicity and training efficiency.

\end{document}